\documentclass[11pt]{article}

\usepackage[preprint]{acl}
\definecolor{gainstrongbg}{RGB}{194,228,204}
\definecolor{gainlightbg}{RGB}{230,243,234}
\definecolor{lossbg}{RGB}{248,228,231}

\newcommand{\poshi}[1]{\begingroup\setlength{\fboxsep}{1pt}\colorbox{gainstrongbg}{#1}\endgroup}
\newcommand{\poslo}[1]{\begingroup\setlength{\fboxsep}{1pt}\colorbox{gainlightbg}{#1}\endgroup}
\newcommand{\negbox}[1]{\begingroup\setlength{\fboxsep}{1pt}\colorbox{lossbg}{#1}\endgroup}

\usepackage{capt-of}
\usepackage{times}
\usepackage{latexsym}
\usepackage{booktabs}
\usepackage{amsmath}
\usepackage{amssymb}
\usepackage{multirow}
\usepackage{makecell}
\usepackage{tabularx}
\newcolumntype{Y}{>{\centering\arraybackslash}X}
\usepackage{array}
\usepackage[table]{xcolor}
\usepackage{svg}
\usepackage{float}
\usepackage[T1]{fontenc}
\usepackage[utf8]{inputenc}

\usepackage{microtype}
\usepackage{marvosym}
\usepackage{footmisc}
\usepackage{inconsolata}

\usepackage{graphicx}

\title{\textsc{DiARC}: Distinguishing Positive and Negative Samples Helps Improving ARC-like Reasoning Ability of Large Language Models}

\author{
  Yuxuan Yang$^{*}$, Feiyang Li$^{*}$, Yile Wang\textsuperscript{\Letter} \\
  College of Computer Science and Software Engineering, Shenzhen University
}

\begin{document}

\maketitle

\DefineFNsymbols*{authsymbols}{*\Letter}
\setfnsymbol{authsymbols}
\renewcommand{\thefootnote}{\fnsymbol{footnote}}
\footnotetext[1]{Equal contribution.}
\footnotetext[2]{Corresponding to \textit{wangyile@szu.edu.cn}.}
\renewcommand{\thefootnote}{\arabic{footnote}}
\setcounter{footnote}{0}

\begin{abstract}
The Abstraction and Reasoning Corpus (ARC;~\citealp{chollet2019measure}) contains tasks that require summarizing patterns from limited grid samples and predicting output grids. Recently, many large language model based approaches have attempted to transform it into a text-based reasoning task. However, methods based on open-source models have generally yielded unsatisfactory results, while those relying on closed-source models are too costly. Current efforts mainly focus on data augmentation, constructing ARC-like data for more comprehensive supervised fine-tuning. In this work, we argue that solving ARC-like problems requires not only \textit{positive} sample supervision but also the ability to improve model reasoning by distinguishing \textit{negative} samples. To this end, we draw on the idea of preference alignment and propose \textsc{DiARC}, a method that constructs preference pairs to enable the model to distinguish between them. Specifically, we propose three ways to construct negative samples, including output-level visual transformations, DSL-level rule inversion, and task-specific rule editing. The resulting negative samples provide informative near-miss alternatives while keeping the observed demonstrations unchanged. Experimental results across multiple ARC-like benchmarks show that \textsc{DiARC} consistently improves performance over baseline models. The code is released at \url{https://github.com/szu-tera/DiARC}.
\end{abstract}

\section{Introduction}

\begin{figure}[t!]
    \centering
    \includegraphics[width=0.98\linewidth]{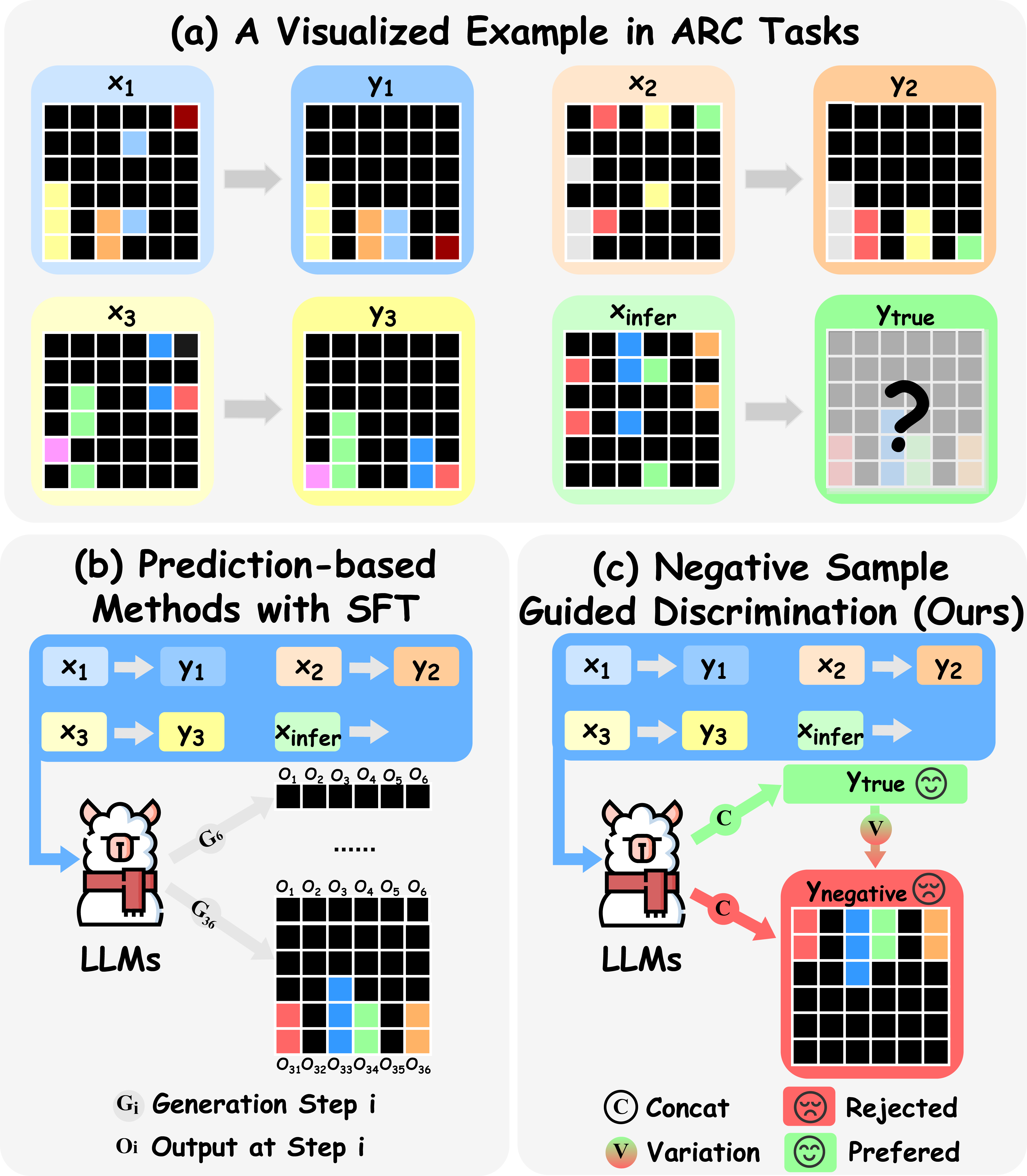}
    \caption{Schematic of ARC tasks and our method. (a) A visualized example of ARC problem. (b) Conventional methods with correct target supervision only. (c) \textsc{DiARC} leverage negative samples and optimize the model with preference alignment objectives.}
    \label{fig:intro}
\end{figure}

Large language models (LLMs) have achieved impressive progress in natural language understanding \citep{yang2024harnessing} and mathematical reasoning \citep{ahn2024mathreasoning}. However, their ability for solving abstract symbolic tasks still have some limitations~\cite{gendron2023large,wang2024speak} and recent work try improving LLMs to solve the typical Abstraction and Reasoning Corpus (ARC;~\citealp{chollet2019measure}), where the model must infer a latent transformation rule from a small number of grid-based input-output examples and apply it to a new test input, as Figure~\ref{fig:intro}(a) shows.

%However, whether they can infer abstract rules from only a few examples and generalize them to new instances remains an open question. Originally proposed as a test of abstract reasoning and generalization \citep{chollet2019measure}, the Abstraction and Reasoning Corpus (ARC) has since become a representative and challenging benchmark for evaluating such abilities in modern foundation models, as illustrated in Figure~\ref{fig:intro}(a). 

%In ARC, a model must infer a latent transformation rule from a small number of grid-based input-output examples and apply it to a new test input. Yet current results remain limited, and stronger frontier LLM-based systems often require substantial inference-time cost to achieve higher ARC performance, underscoring the difficulty of abstract rule induction in ARC \citep{arcprizeleaderboard,vahdati2026arcprogress}.

Yet current results remain limited, and stronger frontier LLM-based systems often require substantial inference-time cost to achieve higher ARC performance, underscoring the difficulty of abstract rule induction in ARC \citep{arcprizeleaderboard,vahdati2026arcprogress}. Recent progress on improving ARC reasoning ability of LLMs has increasingly relied on expanding supervision beyond the original benchmark tasks. \citet{franzen2025product} builds training pipeline around ARC-specific data augmentation using the RE-ARC dataset~\citep{hodel2024addressing}. \citet{wang2025mixture} propose program-based data synthesis for ARC tasks to expose the model to substantially more ARC-style supervision. NVARC~\cite{nvarc2025} pushes this trend further by introducing a multi-stage synthetic data generation pipeline, where LLMs are used to produce puzzle summaries, compose more complex task descriptions, and generate corresponding input-output grid logic in code, yielding a large synthetic corpus for ARC-style training. Although these approaches differ in how additional supervision is constructed, the overall pipelines still rely only on input-output supervision over correct targets, as Figure~\ref{fig:intro}(b) shows.

From the perspective of learning and cognitive science, learning on difficult problems does not arise only from correct examples, but also from identifying and correcting informative errors~\cite{metcalfe2017learning,dieterich2025erroneous}. This idea is also useful for training LLMs. For example, learning from step-level preference signals and erroneous traces can improve long-chain mathematical reasoning performance \citep{lai2024stepdpo,lu2024scdpo}. \citet{legris2025comprehensive} also shows that human solvers in ARC can improve across attempts using feedback. Inspired by them, we propose \textsc{DiARC}, a preference alignment framework for improving ARC reasoning ability through encourage LLMs to distinguish the correct output from rejected alternatives, as an example shown in Figure~\ref{fig:intro}(c).

Specifically, \textsc{DiARC} constructs preference pairs in which the chosen response is the answer to the query input and the rejected response is a negative derived from that task. The negative samples are constructed from three levels: (1) visual modifications applied directly to the output grid to produce candidates that remain plausible while violating the underlying rule; (2) recurring domain-specific language patterns extracted from task-specific transformation programs, yielding reusable program-level rewrites that produce rule-level negatives across tasks; and (3) LLM-assisted edits to each task's transformation rule, constructing task-specific negatives that are as close as possible to the semantic opposite of the original rule. Training on these preference pairs allows to optimize model not only for correctness, but also for discrimination between the correct rule and plausible near-miss alternatives.

We evaluate on six ARC benchmarks including: ARC-AGI-1 \citep{chollet2019measure}, ARC-AGI-2 \citep{chollet2025arcagi2}, MiniARC \citep{kim2022playgrounds}, ConceptARC \citep{moskvichev2023conceptarc}, 1D-ARC \citep{xu2024llm4arc}, and ARCcommunity. Experimental results using three open-source LLMs show that \textsc{DiARC} yields consistent gains across all ARC datasets. By using Qwen3 model, it achieves over 96\% accuracy on ARC-AGI-1, MiniARC, and ConceptARC, outperforming various ARC-specialized models and strong closed-source LLMs.
\begin{figure*}[t]
    \centering
  \includegraphics[width=0.98\linewidth]{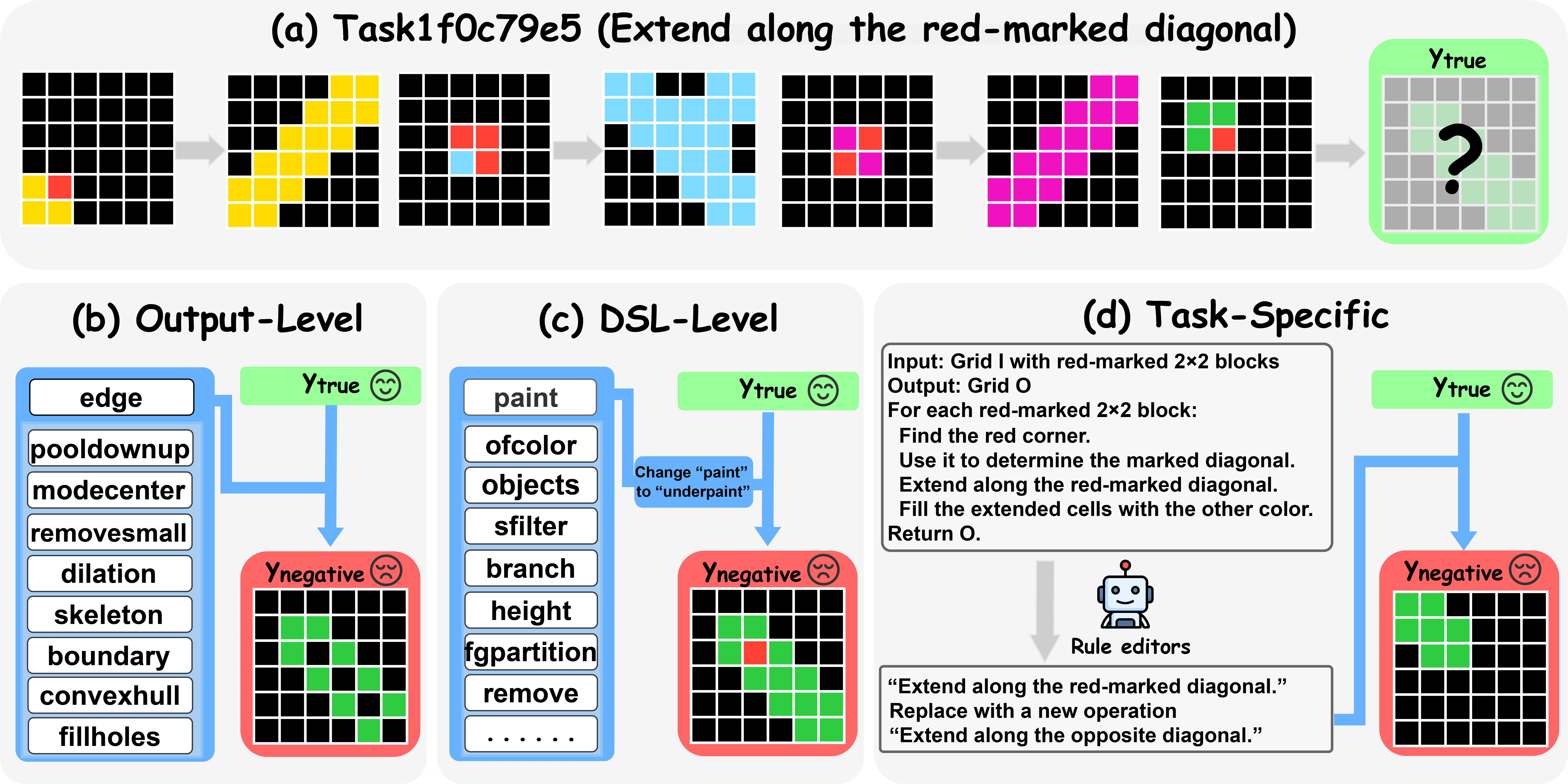}
    \caption{Examples of negative construction strategies for \textsc{DiARC}. (a) Visualization of ARC task 1f0c79e5. (b) Output-level visual transformation ($\S$\ref{sec:method_output}). (c) DSL-level rule inversion  ($\S$\ref{sec:method_dsl}). (d) Task-specific rule editing  ($\S$\ref{sec:method_task}).}
    \label{fig:method_examples}
\end{figure*}

\section{Preliminaries and Motivations}

\paragraph{ARC Task Formulation.}
An ARC-like task $\tau$ consists of a support set $\mathcal{S}_{\tau}=\{(I_i, O_i)\}_{i=1}^{n_\tau}$, a query input grid $q_\tau$, and a target output grid $O_\tau^{*}$. Each grid is transferred to numerical entries in $\{0,\dots,9\}$, where each integer denotes a grid with corresponding color. Each task is governed by a latent task-specific transformation $T_\tau$ such that $O_i = T_\tau(I_i)$ for $i=1,\dots,n_\tau$, and $O_\tau^{*} = T_\tau(q_\tau)$.

The support set $\mathcal{S}_{\tau}$ and query $q_\tau$ are serialized to the input prompt $x_\tau$, and the  target output $O_\tau^{*}$ is serialized to the output $y_\tau$ for LLMs. Thus, each ARC task is converted into a standard sequence-to-sequence example $(x_\tau, y_\tau)$ and can be treated as next grid prediction in a language modeling style.

\paragraph{Supervised Fine-Tuning.}
Given a autoregressive LLMs $\pi_{\theta}$, let $\mathcal{D}_{\mathrm{SFT}}=\{(x_\tau, y_\tau)\}$ denote the training set constructed from serialized ARC tasks, supervised fine-tuning (SFT) optimizes the model by maximizing the conditional likelihood of each token in target sequence given the input prompt:
\begin{equation}
\scalebox{0.84}{$
\mathcal{L}_{\mathrm{SFT}}^\theta
=
-\mathbb{E}_{(x_\tau,y_\tau)\sim\mathcal{D}_{\mathrm{SFT}}}
\left[
\sum_{t=1}^{|y_\tau|}
\log \pi_{\theta}(y_{\tau_t} \mid x_\tau, y_{\tau_{<t}})
\right].
$}
\label{eq:sft_loss}
\end{equation}

This method uses all positive samples, i.e., correct input-output pairs, and requires the model to accurately predict the integer corresponding to the next grid at each prediction step during training, as Figure~\ref{fig:intro}(b) shows.

%\subsection{ARC Prize Inference Pattern}

%The two ARC Prize system lines we follow adopt the same inference pattern rather than relying on a base model alone. Starting from an ARC-specialized model, the pipeline first performs test-time training (TTT) on the support examples of the current task. It then applies task augmentation (Aug) to construct multiple transformed views of the task and generates candidate outputs under these views. Next, Product-of-Experts scoring (PoE) uses the model itself both as a generator and as a discriminator: the model first proposes candidate outputs and then scores them across augmented views, so that candidates more consistently supported by the task receive higher preference. Finally, depth-first search (DFS) replaces standard stochastic sampling with a search-based decoding procedure to explore candidate solutions more effectively.

\paragraph{Our Motivations.}
Our motivation consists of three parts: 1) The output sequences of ARC tasks are often very long, making it difficult to accurately predict at each step from a generative perspective. 2) SFT-based methods focus on local information, whereas ARC tasks are designed with global grid characteristics in mind. 3) SFT-based methods only rely on a large number of positive samples and lack guidance from incorrect outputs. Compared with generation, holistically judging which output is more reasonable is a relatively easy skill to acquire, yet it is a crucial ability for ARC reasoning. Thereby, inspired by preference alignment post-training~\cite{ouyang2022training}, we consider using preference optimization to further enhance the model's ability to solve ARC tasks.

\section{Method}

\subsection{Overview}

%Given a raw ARC task context $c=(\mathcal{S}, q)$, where $\mathcal{S}=\{(I_i,O_i)\}_{i=1}^{n}$ is the support set and $q$ is the query input grid, let $O^{*}$ denote the target output grid for the query. We write $x=\mathrm{ser}(c)$ for the serialized prompt and $y^{*}=\mathrm{ser}(O^{*})$ for the corresponding target sequence. Our goal is to train the model not only to prefer the correct output, but also to reject plausible yet rule-incorrect alternatives.

Given the task with support set $\mathcal{S}_{\tau}$, query $q_\tau$, target $O_\tau^{*}$, the serialized input and output $(x^*, y^*)$, we denote the chosen output by $y^{+}=y^{*}$. We aim to optimize the model not only to prefer the correct output, but also to reject plausible yet rule-incorrect alternatives during post-training. 

To this end, we use a \textbf{negative samples generator} ($\mathcal{NG}$ in short) to construct a rejected output for the query while keeping the support demonstrations unchanged. We denote negative sample $y^{-}=\mathcal{NG}(\mathcal{S}_{\tau}, q_\tau, y^{+})$, where the implementation of $\mathcal{NG}$  depends on the negative-construction strategy. We then obtain a preference dataset $\mathcal{D}_{\text{pref}}=\{(x,y^{+},y^{-})\}$, where $y^{+}$ is the chosen output and $y^{-}$ is the rejected output. To construct informative rejected alternatives, we generate negatives at three levels ranging from visual modifications of the output grid to program-level rule reversal, including output-level visual transformation ($\S$\ref{sec:method_output}), DSL-level rule inversion ($\S$\ref{sec:method_dsl}), task-specific rule editing ($\S$\ref{sec:method_task}), and we introduce them as follows.

%The first strategy operates directly in the output space through visual transformations, motivated by the strongly visual nature of ARC tasks \citep{hu2025arcvision}. In addition, ARC has also inspired a line of work based on explicit programmatic or DSL-based representations \citep{hodel2024addressing,rocha2025ilparc}. Motivated by this direction, our second strategy performs reusable DSL-level rule inversion over recurring patterns extracted from task-specific transformation programs, and our third strategy constructs task-specific negatives through LLM-assisted edits to each task's transformation rule. 

%These three strategies operate at different granularities, ranging from visual modifications of the output grid to program-level rule reversal. The resulting preference pairs are then used to train \textsc{DiARC} with a pairwise preference objective, instantiated here with DPO. 

%Figure~\ref{fig:method_examples} presents a representative ARC task and example negatives produced by the three strategies, and Figure~\ref{fig:method_pipeline} shows the overall preference-construction and DPO training pipeline.

\subsection{Output-Level Visual Transformations}
\label{sec:method_output}
The first strategy $\mathcal{NG}_{\mathrm{Vis}}$ constructs negatives directly in the output space through visual transformations of the output grid, motivated by the strongly visual nature of ARC tasks \citep{hu2025arcvision}, as illustrated in Figure~\ref{fig:method_examples}(b).
The key idea is to generate candidate outputs that remain visually and structurally plausible while deviating from the correct transformation rule. We use four families of atomic operations, yielding 28 transformations in total.

The first family, \textit{grid-block} operators, contains three transformations: \texttt{pooldownup}, which applies $2\times2$ mode pooling followed by upsampling; \texttt{modecenter}, which replaces only the center cell of each $3\times3$ block with the block mode; and \texttt{removesmall}, which removes connected components smaller than a preset size threshold. These operations preserve coarse layout while damaging fine-grained structure.

The second family, \textit{rigid shifts}, contains 16 transformations. We shift the full grid content in eight directions $\{\mathcal{U},\mathcal{D},\mathcal{L},\mathcal{R},\mathcal{UL},\mathcal{UR},\mathcal{DL},\mathcal{DR}\}$ and use two magnitudes, 10\% and 30\%, with newly exposed cells filled by the background color. This produces spatially coherent but misaligned outputs.

The third family consists of seven morphological and geometry-related operators: \texttt{erosion}, \texttt{dilation}, \texttt{skeleton}, \texttt{edge}, \texttt{boundary}, \texttt{convexhull}, and \texttt{fillholes}. Intuitively, these operators thin, expand, outline, simplify, or complete objects while keeping the overall spatial layout largely unchanged.

The fourth family consists of two local perturbation operators: \texttt{noise5}, which injects 5\% salt-and-pepper noise, and \texttt{swap10}, which swaps 10\% of boundary pixels with neighboring pixels of different colors. These operations introduce localized appearance and boundary inconsistencies while largely preserving the global arrangement of objects in the grid.

\subsection{DSL-Level Rule Inversion}
\label{sec:method_dsl}
ARC tasks are represented in a higher-level domain-specific language (DSL) as executable transformation programs built from predefined primitives. Prior works represent solutions as executable programs and derives additional supervision through program mutation or decomposition \citep{butt2024codeit,simpson2026decomposing}. Based on them, the second strategy $\mathcal{NG}_{\mathrm{DSL}}$ construct negatives at the program level by inverting recurring local rule patterns in task-specific transformation programs, as shown in Figure~\ref{fig:method_examples}(c). 

We first analyze a pool of task-specific programs that map input grids to output grids, identify recurring local DSL patterns across tasks, and retain those with relatively stable semantics. To obtain a reusable pattern set, we apply a greedy coverage criterion: at each step, we select the candidate pattern whose rewrite covers the largest number of previously uncovered tasks. For each retained pattern $m$, we then define a unified program-level rewrite $\phi_m$ that acts as a semantic reversal of the original local rule. 

Given a task program $p$ and query input $q$, the original program produces the chosen output $y^{+}=p(q)$, while the rewritten program produces the rejected output $y^{-}=\phi_m(p)(q)$. The resulting negative is therefore rule-incorrect due to a localized semantic inversion, while remaining close to the original task logic.

\subsection{Task-Specific Rule Editing}
\label{sec:method_task}
Motivated by program refinement and explicit rule-guided code editing based works~\cite{pourcel2025soar,li2025editlord}, the third strategy $\mathcal{NG}_{\mathrm{Edit}}$ constructs negatives at the level of individual tasks by using LLM-assisted edits to each task's transformation rule, as Figure~\ref{fig:method_examples}(d) shows.

Concretely, for a task-specific transformation program, we use GPT-5.4~\cite{gpt54} to assist rule editing and construct an alternative rule that is as close as possible to the semantic opposite of the original rule. When a clean reversal is not feasible, GPT-5.4 instead generates a task-specific alternative rule that stays close to the original transformation logic but leads to a different transformation behavior. The resulting negative therefore stays close to the original solution in overall structure and execution pattern, while deviating in the task-relevant rule itself.

\begin{figure}[t!]
    \centering
    \includegraphics[width=1\linewidth]{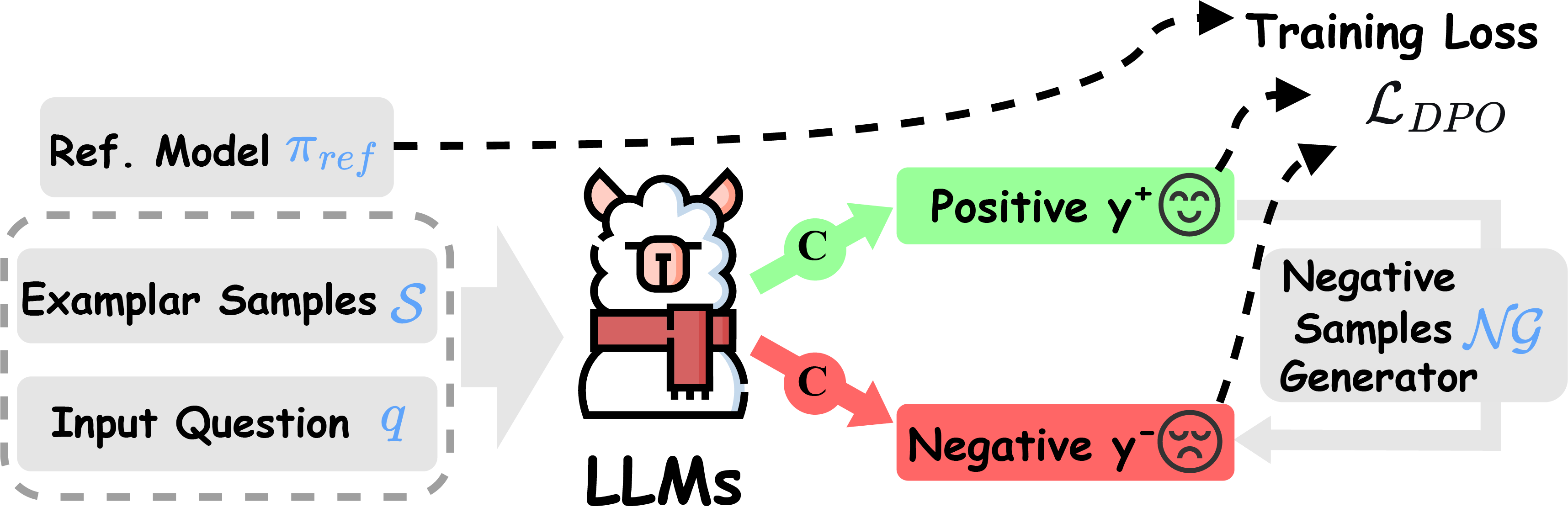}
   \caption{Preference-learning pipeline. Given an ARC task, we serialize the context, construct a chosen output and a rejected output, and optimize the model with DPO.}
    \label{fig:method_pipeline}
\end{figure}

\subsection{Preference Optimization Pipeline}
After obtaining the negative samples and the constructed preference data, we train the model with direct preference optimization (DPO; \citealp{rafailov2023direct}). Let $\pi_{\theta}$ denote the policy model and $\pi_{\mathrm{ref}}$ the reference model. For each preference triple $(x,y^{+},y^{-})$, DPO encourages the model to assign higher relative likelihood to the chosen output than to the rejected output:
\begin{equation}
\scalebox{0.8}{$
\begin{aligned}
\mathcal{L}_{\mathrm{DPO}}^\theta
=
-\mathbb{E}_{(x,y^{+},y^{-})\sim\mathcal{D}_{\mathrm{pref}}}
\log \sigma \Bigg(
\beta &\log \frac{\pi_{\theta}(y^{+}\mid x)}{\pi_{\mathrm{ref}}(y^{+}\mid x)}
\\
-
\beta &\log \frac{\pi_{\theta}(y^{-}\mid x)}{\pi_{\mathrm{ref}}(y^{-}\mid x)}
\Bigg),
\end{aligned}
$}
\label{eq:dpo_pair}
\end{equation}
where $\beta$ is a temperature parameter. In practice, we initialize $\pi_{\theta}$ from $\pi_{\mathrm{SFT}}$ and use a frozen copy of $\pi_{\mathrm{SFT}}$ as the reference model $\pi_{\mathrm{ref}}$. Unlike reinforcement learning methods that require an explicit reward model, DPO directly optimizes pairwise preferences between correct and incorrect outputs. This makes it particularly suitable for ARC, where supervision is sparse and many errors are better characterized as plausible but rule-incorrect alternatives rather than compressed into a single reward score. Figure~\ref{fig:method_pipeline} shows the overall pipeline of preference data construction and DPO training.

\section{Experiments}
\label{sec:experiments}

\subsection{Experimental Setup}
\label{subsec:exp_setup}

\noindent\textbf{Dataset and Evaluation.}
We evaluate on six benchmarks that cover the original ARC setting, its harder successor, and four related ARC-style transfer or additional benchmarks. \textbf{ARC-AGI-1}~\cite{chollet2019measure} is the original ARC benchmark for few-shot abstract rule induction from grid-based input-output examples, and we use its 400-task public evaluation set. \textbf{ARC-AGI-2}~\cite{chollet2025arcagi2} is a harder successor that preserves the same input-output format while providing a more challenging test of abstract reasoning and problem solving, and we use its 120-task public evaluation set. \textbf{MiniARC}~\cite{kim2022playgrounds} is a compact 5$\times$5 ARC-style benchmark designed to reduce variation in grid size and focus more directly on abductive reasoning. \textbf{ConceptARC}~\cite{moskvichev2023conceptarc} organizes ARC-style tasks into concept groups in order to evaluate abstraction and generalization over recurring spatial and semantic concepts. \textbf{1D-ARC}~\cite{xu2024llm4arc} is a one-dimensional ARC-style benchmark that simplifies the original two-dimensional grid setting into one-dimensional transformation tasks while preserving the core requirement of abstract rule induction. \textbf{ARCcommunity}\footnote{\url{https://github.com/arc-community/arc}} is a collection of community-created ARC-style tasks that extends evaluation beyond the original benchmark distribution and provides an additional test of generalization to diverse human-designed problems.

For the standard evaluation on ARC-AGI-1 and ARC-AGI-2, each test input allows two candidate outputs and is counted as correct if either exactly matches the ground-truth grid. If a task contains multiple test inputs, the task score is the average over its test inputs. For MiniARC, ConceptARC, 1D-ARC, and ARCcommunity, we report single-response exact-match accuracy.

\noindent\textbf{Models.}
We use LLMs across three families including Llama-3.2-3B~\cite{grattafiori2024llama}, Mistral-NeMo-Minitron-8B~\cite{sreenivas2024llm}, and Qwen3-4B~\cite{yang2025qwen3}. During training, we optimized their vocabulary and the effective fine-tuned parameter size is slightly reduced.

%The first two are our reproductions from the ARChitects/Product-of-Experts line of work \citep{thearchitects2024,franzen2025product}, i.e., the first-place open-source model line from ARC Prize 2024. For \textbf{Qwen3-4B}, we directly use the official released parameters from \textbf{NVARC} \citep{nvarc2025}, the first-place ARC Prize 2025 system.

\begin{table}[t!]
\centering
\scalebox{0.6}{
\begin{tabular}{lrrr}
\toprule
\textbf{Models} & \#\textbf{Params} & \textbf{ARC-1} & \textbf{ARC-2} \\
\midrule
\specialrule{\lightrulewidth}{0pt}{0pt}
\rowcolor{gray!15}
\multicolumn{4}{l}{\textit{Closed-source Large Language Models}}\\
Opus 4.5~\citep{opus45}$^\dag$  & N/A & 40.0 & 7.8 \\
Kimi K2.5~\citep{team2026kimi}$^\dag$  & 1T & 65.3 & 11.8 \\
GPT-5.4~\citep{gpt54}$^\dag$ & N/A & 68.2 & 29.2 \\
Grok-4~\citep{grok4}$^\ddag$ & 1.7T & 79.6 & 29.4 \\
\midrule
\specialrule{\lightrulewidth}{0pt}{0pt}
\rowcolor{gray!15}
\multicolumn{4}{l}{\textit{ARC-specialized Recursive Models}}\\
HRM~\citep{wang2025hrm} & 27M & 40.3 & 5.0 \\
GRAM~\citep{baek2026generativerecursivereasoning} & 10.9M & 52.0 & 11.1 \\
\midrule
\specialrule{\lightrulewidth}{0pt}{0pt}
\rowcolor{gray!15}
\multicolumn{4}{l}{\textit{ARC-specialized Vision Models}}\\
VARC~\citep{hu2025arcvision} & 18M & 54.5 & 8.3 \\
VARC (ensemble)~\citep{hu2025arcvision} & 73M & 60.4 & 11.1 \\

\midrule
\specialrule{\lightrulewidth}{0pt}{0pt}
\rowcolor{gray!15}
\multicolumn{4}{l}{\textit{Open-source Large Language Models}}\\
Llama-3.2~\cite{grattafiori2024llama}  & 3B & 0.50 & 0.00 \\
Llama-3.2 + SFT only~\citep{franzen2025product} & 2.6B & 59.38 & 4.44 \\
\textbf{Llama-3.2 + \textsc{DiARC} (Ours)} & 2.6B & \textbf{61.50} & \textbf{6.53} \\
\midrule
Minitron~\cite{sreenivas2024llm} & 8B & 1.12 & 0.00 \\
Minitron + SFT only~\citep{franzen2025product} & 6.9B & 71.25 & 9.86 \\
\textbf{Minitron + \textsc{DiARC} (Ours)} & 6.9B & \textbf{72.38} & \textbf{10.28} \\
\midrule
Qwen3~\cite{yang2025qwen3}  & 4B & 0.63 & 0.00 \\
Qwen3 + SFT only~\citep{nvarc2025} & 3.6B & 96.25 & 30.83 \\
\textbf{Qwen3 + \textsc{DiARC} (Ours)} & 3.6B & \textbf{97.00} & \textbf{33.06} \\
\midrule
\specialrule{\lightrulewidth}{0pt}{0pt}
\rowcolor{gray!15}
\multicolumn{4}{l}{\textit{Human Performance}}\\
Human panel (average)~\citep{arcprize2025launch} & -- & 64.20 & 60.00 \\
\bottomrule
\end{tabular}}
\caption{Results on ARC-AGI-1 and ARC-AGI-2 benchmarks. $\dag$ : Reported in leaderboard of \citet{arcprizeleaderboard}. $\ddag$: Reported in \citet{berman2025highest}.}
\label{tab:arcagi_benchmarks}
\end{table} 

\begin{table}[t!]
\centering
\scalebox{0.49}{
\begin{tabular}{lrrrrr}
\toprule
\textbf{Models} & \#\textbf{Params} & \textbf{Mini} & \textbf{Concept} & \textbf{1D} & \textbf{Comm.}\\
\midrule
\specialrule{\lightrulewidth}{0pt}{0pt}
\rowcolor{gray!15}
\multicolumn{6}{l}{\textit{Closed-source Large Language Models}}\\
o1-mini~\citep{o1_mini}$^\dag$ & N/A & 30.20 & -- & -- & -- \\
DeepSeek-R1~\citep{ds_r1}$^\dag$ & 671B & 28.86 & -- & -- & -- \\
%DeepSeek-V3~\citep{ds_r3}$^\dag$ & 671B & 27.52 & -- & 69.70 & -- \\
GPT-4~\citep{gpt4}$^\ddag$ & N/A & 24.10 & 24.0 & 52.00 & -- \\
Qwen-2.5~\citep{qwen25}$^\ddag$ & 7B & 22.00 & 26.0 & 55.00 & -- \\
Claude Sonnet 4~\citep{sonnet4}$^\natural$ & N/A & -- & 60.2 & -- & -- \\
Gemini 2.5 Pro~\citep{gemini25}$^\natural$ & N/A & -- & 66.0 & -- & -- \\
o4-mini~\citep{o4_mini}$^\natural$ & N/A & -- & 70.8 & -- & -- \\

\midrule
\specialrule{\lightrulewidth}{0pt}{0pt}
\rowcolor{gray!15}
\multicolumn{6}{l}{\textit{Closed-source Large Vision Models}}\\
o4-mini (visual)~\citep{o4_mini}$^\natural$ & N/A & -- & 8.1 & -- & -- \\

\midrule
\specialrule{\lightrulewidth}{0pt}{0pt}
\rowcolor{gray!15}
\multicolumn{6}{l}{\textit{Open-source Large Language Models}}\\
Llama-3.2~\cite{grattafiori2024llama} & 3B & 1.34 & 1.04 & 0.11 & 2.17 \\
Llama-3.2 + SFT only~\citep{franzen2025product} & 2.6B & 70.47 & 58.13 & 51.72 & 23.91 \\
\textbf{Llama-3.2 + \textsc{DiARC} (Ours)} & 2.6B & \textbf{73.15} & \textbf{59.17} & \textbf{52.39} & \textbf{30.43} \\

\midrule
Minitron~\cite{sreenivas2024llm} & 8B & 6.04 & 5.83 & 3.55 & 6.52 \\
Minitron + SFT only~\citep{franzen2025product} & 6.9B & 74.50 & 60.42 & 57.05 & 34.78 \\
\textbf{Minitron + \textsc{DiARC} (Ours)} & 6.9B & \textbf{76.51} & \textbf{62.71} & \textbf{58.82} & \textbf{43.48} \\
\midrule
Qwen3~\cite{yang2025qwen3} & 4B & 0.67 & 1.25 & 0.11 & 0.00 \\
Qwen3 + SFT only~\citep{nvarc2025} & 3.6B & 95.30 & 96.46 & 42.84 & 58.70 \\
\textbf{Qwen3 + \textsc{DiARC} (Ours)} & 3.6B & \textbf{97.32} & \textbf{97.29} & \textbf{45.95} & \textbf{63.04} \\
\specialrule{\lightrulewidth}{0pt}{0pt}
\rowcolor{gray!15}
\multicolumn{6}{l}{\textit{Human Performance}}\\
Human (pass@1)~\citep{beger2025humanlike} & -- & -- & 73.00 & -- & -- \\

\bottomrule
\end{tabular}}
\caption{Results on MiniARC, ConceptARC, 1D-ARC, and ARCcommunity benchmarks. $\dag$ : Reported in \citet{zheng2025cursecot}. $\ddag$: Reported in \citet{wang2025mixture}. $\natural$: Reported in \citet{beger2025humanlike}.}
\label{tab:transfer_benchmarks}
\end{table}

\begin{table*}[t!]
\centering
\scriptsize
\setlength{\tabcolsep}{4.2pt}
\renewcommand{\arraystretch}{1.08}
\begin{tabular*}{\textwidth}{@{\extracolsep{\fill}}llrrrrr@{}}
\toprule
\textbf{Benchmarks} & \textbf{Models} & \textbf{SFT Init.} & \textbf{Grid-Block} & \textbf{Morphology} & \textbf{Random Perturb} & \textbf{Rigid Shift} \\
\midrule

\multirow{3}{*}{ARC-AGI-1}
& Llama-3.2-3B & 59.38 & 59.88 {\scriptsize\poslo{(+0.50)}} & 60.75 {\scriptsize\poshi{(+1.37)}} & 61.50 {\scriptsize\poshi{(+2.12)}} & 59.75 {\scriptsize\poslo{(+0.37)}} \\
& Minitron-8B  & 71.25 & 71.88 {\scriptsize\poslo{(+0.63)}} & 71.88 {\scriptsize\poslo{(+0.63)}} & 71.38 {\scriptsize\poslo{(+0.13)}} & 71.38 {\scriptsize\poslo{(+0.13)}} \\
& Qwen3-4B     & 96.25 & 96.75 {\scriptsize\poslo{(+0.50)}} & 97.00 {\scriptsize\poslo{(+0.75)}} & 96.50 {\scriptsize\poslo{(+0.25)}} & 96.75 {\scriptsize\poslo{(+0.50)}} \\
\midrule

\multirow{3}{*}{ARC-AGI-2}
& Llama-3.2-3B & 4.44  & 6.11 {\scriptsize\poshi{(+1.67)}} & 5.69 {\scriptsize\poshi{(+1.25)}} & 6.53 {\scriptsize\poshi{(+2.09)}} & 5.69 {\scriptsize\poshi{(+1.25)}} \\
& Minitron-8B  & 9.86  & 9.31 {\scriptsize\negbox{(-0.55)}} & 10.28 {\scriptsize\poslo{(+0.42)}} & 9.17 {\scriptsize\negbox{(-0.69)}} & 9.03 {\scriptsize\negbox{(-0.83)}} \\
& Qwen3-4B     & 30.83 & 31.39 {\scriptsize\poslo{(+0.56)}} & 31.81 {\scriptsize\poslo{(+0.98)}} & 32.64 {\scriptsize\poshi{(+1.81)}} & 33.06 {\scriptsize\poshi{(+2.23)}} \\
\midrule

\multirow{3}{*}{MiniARC}
& Llama-3.2-3B & 70.47 & 71.14 {\scriptsize\poslo{(+0.67)}} & 73.15 {\scriptsize\poshi{(+2.68)}} & 72.48 {\scriptsize\poshi{(+2.01)}} & 72.48 {\scriptsize\poshi{(+2.01)}} \\
& Minitron-8B  & 74.50 & 75.17 {\scriptsize\poslo{(+0.67)}} & 76.51 {\scriptsize\poshi{(+2.01)}} & 75.17 {\scriptsize\poslo{(+0.67)}} & 76.51 {\scriptsize\poshi{(+2.01)}} \\
& Qwen3-4B     & 95.30 & 97.32 {\scriptsize\poshi{(+2.02)}} & 97.32 {\scriptsize\poshi{(+2.02)}} & 97.32 {\scriptsize\poshi{(+2.02)}} & 96.64 {\scriptsize\poshi{(+1.34)}} \\
\midrule

\multirow{3}{*}{ConceptARC}
& Llama-3.2-3B & 58.13 & 58.75 {\scriptsize\poslo{(+0.62)}} & 59.17 {\scriptsize\poshi{(+1.04)}} & 59.17 {\scriptsize\poshi{(+1.04)}} & 58.13 {\scriptsize\poslo{(+0.00)}} \\
& Minitron-8B  & 60.42 & 62.71 {\scriptsize\poshi{(+2.29)}} & 62.29 {\scriptsize\poshi{(+1.87)}} & 62.71 {\scriptsize\poshi{(+2.29)}} & 62.08 {\scriptsize\poshi{(+1.66)}} \\
& Qwen3-4B     & 96.46 & 96.88 {\scriptsize\poslo{(+0.42)}} & 96.88 {\scriptsize\poslo{(+0.42)}} & 96.67 {\scriptsize\poslo{(+0.21)}} & 97.29 {\scriptsize\poslo{(+0.83)}} \\
\midrule

\multirow{3}{*}{1D-ARC}
& Llama-3.2-3B & 51.72 & -- & -- & 52.39 {\scriptsize\poslo{(+0.67)}} & 52.28 {\scriptsize\poslo{(+0.56)}} \\
& Minitron-8B  & 57.05 & -- & -- & 58.82 {\scriptsize\poshi{(+1.77)}} & 58.16  {\scriptsize\poshi{(+1.11)}} \\
& Qwen3-4B     & 42.84 & -- & -- & 45.95 {\scriptsize\poshi{(+3.11)}} & 45.17 {\scriptsize\poshi{(+2.33)}} \\
\midrule

\multirow{3}{*}{ARCcommunity}
& Llama-3.2-3B & 23.91 & 30.43 {\scriptsize\poshi{(+6.52)}} & 26.09 {\scriptsize\poshi{(+2.18)}} & 26.09 {\scriptsize\poshi{(+2.18)}} & 26.09 {\scriptsize\poshi{(+2.18)}} \\
& Minitron-8B  & 34.78 & 43.48 {\scriptsize\poshi{(+8.70)}} & 43.48 {\scriptsize\poshi{(+8.70)}} & 43.48 {\scriptsize\poshi{(+8.70)}} & 43.48 {\scriptsize\poshi{(+8.70)}} \\
& Qwen3-4B     & 58.70 & 60.87 {\scriptsize\poshi{(+2.17)}} & 63.04 {\scriptsize\poshi{(+4.34)}} & 58.70 {\scriptsize\poslo{(+0.00)}} & 63.04 {\scriptsize\poshi{(+4.34)}} \\
\midrule

Avg. & -- & 56.31 & 58.14 {\scriptsize\poshi{(+1.83)}} & 58.36 {\scriptsize\poshi{(+2.04)}} & 57.97 {\scriptsize\poshi{(+1.66)}} & 58.09 {\scriptsize\poshi{(+1.78)}} \\
\bottomrule
\end{tabular*}
\caption{Results of output-level negative construction families across six benchmarks. Values are formatted as result (\poshi{gain $>$ 1.00}, \poslo{gain $\leq$ 1.00}, or \negbox{loss}) relative to the corresponding SFT initialization model. For 1D-ARC, only random perturb and rigid shift operation are applicable, so results for grid-block and morphology are not reported.}
\label{tab:output_level_family_summary}
\end{table*}\

\noindent\textbf{Preference Data Construction.}
For ARC-AGI-1, all three negative-construction strategies are built on top of RE-ARC~\citep{hodel2024addressing}, which provides task-specific generators for the 400 original ARC training tasks. For \textit{output-level visual transformation}, we first use these generators to sample ARC tasks and then apply output-space transformations to construct rejected outputs. For \textit{DSL-level rule inversion} and \textit{task-specific rule editing}, we operate on the corresponding task-specific transformation programs to generate chosen/rejected pairs. For the remaining five benchmarks, RE-ARC-style per-task generators are unavailable. We therefore use only the \textit{output-level visual transformation} strategy for preference construction. For ARC-AGI-2, preference data is constructed from the released NVARC-full~\cite{nvarc2025} resources. 

For MiniARC, ConceptARC, 1D-ARC, and ARCcommunity, we use their official training sets to construct preference data with the same ARC-oriented negative-construction pipeline. Across all six benchmarks, we keep the overall preference-learning setup as consistent as possible: the original support demonstrations and query input are always preserved, and only the rejected output is varied according to the available negative-construction strategy. This design helps isolate the contribution of \textsc{DiARC} from other sources of variation and makes cross-benchmark and cross-backbone comparisons easier to interpret.

Detailed statistics of the constructed preference datasets used in these experiments are provided in Appendix~\ref{app:data_statistics}.

\noindent\textbf{Training Details.}
We train all models with a shared setup including learning rate $1\times10^{-6}$, effective batch size 8, cosine learning schedule, warmup ratio 0.1, maximum sequence length 4096, and maximum prompt length 3584. LoRA~\cite{hu2022lora} fine-tuning is used and the model-specific settings are summarized in Appendix~\ref{app:training_details}. All experiments are conducted on a single NVIDIA L40 GPU.

\subsection{Main Results}
\label{subsec:main_results}

Tables~\ref{tab:arcagi_benchmarks} and~\ref{tab:transfer_benchmarks} show the results on all six ARC-style benchmarks, Table~\ref{tab:output_level_family_summary} breaks down the results with different output-level negative families, and Table~\ref{tab:rule_level_arc1} compares the DSL-level inversion and task-specific editing on ARC-AGI-1.

\paragraph{\textsc{DiARC} brings broad gains across backbones and benchmarks.}
As shown in Tables~\ref{tab:arcagi_benchmarks} and~\ref{tab:transfer_benchmarks}, \textsc{DiARC} improves the performance of SFT baselines on all six benchmarks for Llama-3.2-3B, Qwen3-4B, and Minitron-8B, with an average absolute gain of 2.48 points across the 18 model--benchmark pairs. The gains are not limited to weaker SFT baselines but also strong ones like Qwen3-4B. The gains also do not scale monotonically with model size: all three backbones benefit from \textsc{DiARC}, with average improvements of 2.52, 2.21, and 2.72 points for Llama-3.2-3B, Qwen3-4B, and Minitron-8B, respectively. When using the Qwen3 backbone model, \textsc{DiARC} achieves an accuracy of over 96 on ARC-AGI-1, MiniARC, and ConceptARC, surpassing both closed-source models and existing open-source model methods, demonstrating its effectiveness.

\paragraph{Useful preference signals can be constructed at multiple levels.}
Table~\ref{tab:output_level_family_summary} shows that the four output-level families improve over the initialization models on average, with morphology giving the strongest mean performance, although the best family varies across datasets and backbones. One possible reason is that ARC is built around a set of human-like priors, especially objectness as well as basic geometry and topology \citep{chollet2019measure}. Compared with other perturbations, morphology-based negatives more directly alter object boundaries, connectivity, and other structural properties while preserving the object layout. This makes them more likely to stay within the kind of object-centric transformation space that ARC is intended to test, and therefore more likely to provide informative near-miss alternatives. At the same time, the most effective negative family differs across backbones, which may be related to differences in the models' underlying capabilities, as they may develop different inductive biases and characteristic error patterns, and thus benefit from different forms of hard negative supervision.

Table~\ref{tab:rule_level_arc1} shows that both DSL-level inversion and task-specific editing provide gains on ARC-AGI-1. Under task-specific editing, however, we observe a temporary performance drop when around 200 tasks are edited, followed by a recovery as the number of edited tasks increases. This is because LLM-assisted editing tends to map negative samples into a few recurring task types, such as set/extraction operations, compressed summaries, and bounding-box or contour extraction. When the edited set is relatively small, these recurring patterns can dominate the training signal and make the model biased toward a narrow set of negative constructions. As the edited set becomes larger and more diverse, this bias is diluted, allowing the model to better capture the intended contrast between positive and negative rules. Taken together, results suggest that the benefit of \textsc{DiARC} does not depend on a single negative-construction recipe.

\begin{table}[t!]
\centering
\scalebox{0.72}{
\begin{tabular}{lccccc}
\toprule
\multirow{2}{*}{\textbf{Models}} & \multirow{2}{*}{\textbf{SFT Init.}} & \multirow{2}{*}{\textbf{DSL-Level}} & \multicolumn{3}{c}{\textbf{Task-Specific Edit}} \\
\cmidrule(lr){4-6}
 &  &  & {100} & {200} & {400} \\
\midrule
Llama-3.2-3B & 59.38 & \textbf{60.25} & 60.12 & 59.13 & 59.62 \\
Minitron-8B  & 71.25 & \textbf{72.38} & 71.88 & 71.38 & 72.00 \\
Qwen3-4B     & 96.25 & 96.75 & \textbf{97.00} & 96.50 & \textbf{97.00} \\
\bottomrule
\end{tabular}}
\caption{Comparison between DSL-level inversion and task-specific editing on ARC-AGI-1. For task-specific editing, we report results under three data settings of 100, 200, and 400 edited tasks, respectively.}
\label{tab:rule_level_arc1}
\end{table}

\section{Analysis}
\label{sec:analysis}
%We further analyze the effectiveness of \textsc{DiARC} by examining whether the gains arise from improved candidate generation, improved candidate discrimination, or both ($\S$\ref{subsec:gen_disc}), investigating the compatibility with existing test-time scaling techniques ($\S$\ref{subsec:compatibility}), and showing a qualitative case study ($\S$\ref{subsec:case}).

\subsection{Impacts of Generation and Discrimination}
\label{subsec:gen_disc}

Our ARC system first generate candidate solutions and then use product-of-experts scoring to select the final answer. \textsc{DiARC} can therefore improve performance through two different channels: it may increase the number of correct candidate solutions produced by the model, or it may improve the model's ability to assign higher preference to correct candidates that are already generated.

\begin{table}[t!]
\centering
\scalebox{0.62}{
\begin{tabular}{lccc}
\toprule
\textbf{Benchmarks} & \textbf{Mean Gen. Gain} & \textbf{Mean Disc. Gain} & \textbf{Mean $\Delta$ Solved} \\
\midrule
ARC-AGI-1    & +1.56 & +1.94 & +3.50 \\
ARC-AGI-2    & +0.75 & +0.50 & +1.25 \\
MiniARC      & +1.17 & +1.33 & +2.50 \\
ConceptARC   & +1.42 & +3.67 & +5.09 \\
1D-ARC       & +7.83 & +6.50& +14.33\\
ARCcommunity & +0.42 & +0.71 & +1.13 \\
\midrule
All Average          & +2.19 & +2.44& +4.63\\
\bottomrule
\end{tabular}%
}
\caption{Impacts of generation and discrimination after \textsc{DiARC} training across six benchmarks. Mean $\Delta$ solved denotes the average increase in the number of correctly solved items relative to the corresponding base model.}
\label{tab:gen_disc_summary}
\end{table}

\begin{table}[t!]
\centering
\scalebox{0.64}{
\begin{tabular}{lccrrrr}
\toprule
\textbf{Models} & \textbf{Variants} & \textbf{Base} & \textbf{+ TTT} & \textbf{+ Aug} & \textbf{+ PoE} & \textbf{+ DFS} \\
\midrule
\multirow{2}{*}{Llama-3.2-3B}
& Init. & 12.37 & 34.62 & 47.88 & 52.12 & 59.38 \\
& \textbf{+ \textsc{DiARC}} & \textbf{12.62} & \textbf{35.00} & \textbf{53.62} & \textbf{53.62} & \textbf{61.50} \\
\midrule
\multirow{2}{*}{Minitron-8B}
& Init. & 17.12 & 41.12& 60.50 & 63.38 & 71.25 \\
& \textbf{+ \textsc{DiARC}} & \textbf{17.37} & \textbf{41.75}& \textbf{61.12} & \textbf{64.37} & \textbf{72.38} \\
\midrule
\multirow{2}{*}{Qwen3-4B}
& Init. & 23.50  & 31.25& 61.25& 63.75& 96.25\\
& \textbf{+ \textsc{DiARC}} & \textbf{23.50}& \textbf{31.87}& \textbf{62.00}& 

\textbf{64.62}& \textbf{97.00}\\
\bottomrule
\end{tabular}}
\caption{Compatibility of \textsc{DiARC} with the test-time scaling techniques. We replace the base model with its \textsc{DiARC}-enhanced counterpart while keeping the remaining inference pipeline unchanged for comparison.}
\label{tab:poe_compatibility}
\end{table}

To distinguish these effects, we define two gain quantities. Let \textit{generation gain} be the number of correct candidates generated after \textsc{DiARC} training but not generated by the base model, minus the number of correct candidates generated by the base model but no longer generated after \textsc{DiARC} training. Let \textit{discrimination gain} be the number of correct candidates that are generated in both settings, not selected by the base model, but selected after \textsc{DiARC} training, minus the number of correct candidates that are generated in both settings, selected by the base model, but not selected after \textsc{DiARC} training. Positive {generation gain} therefore indicates a net gain in candidate generation, while positive {discrimination gain} indicates a net gain in candidate selection.

Table~\ref{tab:gen_disc_summary} summarizes these effects across six benchmarks. Overall, both quantities are positive on average: the mean {generation gain} is $+2.19$ and the mean {discrimination gain} is $+2.44$, while the mean improvement in solved items is $+4.63$ under the benchmark-specific evaluation metric. These results suggest that the gains from \textsc{DiARC} arise from both stages of the pipeline: the model produces more correct candidates and also becomes better at assigning higher preference to correct candidates once they are generated.

\subsection{Compatibility with Test-Time Scaling}
\label{subsec:compatibility}
Different test-time scaling (TTS) methods have been used to improve reasoning performance of LLMs. We further analyze whether \textsc{DiARC} is compatible with different techniques. Following \citet{franzen2025product}, we combine base and \textsc{DiARC}-enhanced ones with test-time training (TTT), task augmentation (Aug), reranking (PoE, product-of-experts scoring), and searching (DFS, depth-first search) methods, comparing how improvements in the underlying models propagate to final accuracy. 

Table~\ref{tab:poe_compatibility} shows the results. Across different TTS techniques, \textsc{DiARC}-enhanced model remain compatible and preserve competitive performance relative to their corresponding base models. This suggests that \textsc{DiARC} is not limited to direct model evaluation, but can also be integrated into advanced test-time scaling pipeline for model improvement.

\subsection{Case Study}
\label{subsec:case}

\begin{figure}[t!]
    \centering
    \includegraphics[width=1\linewidth]{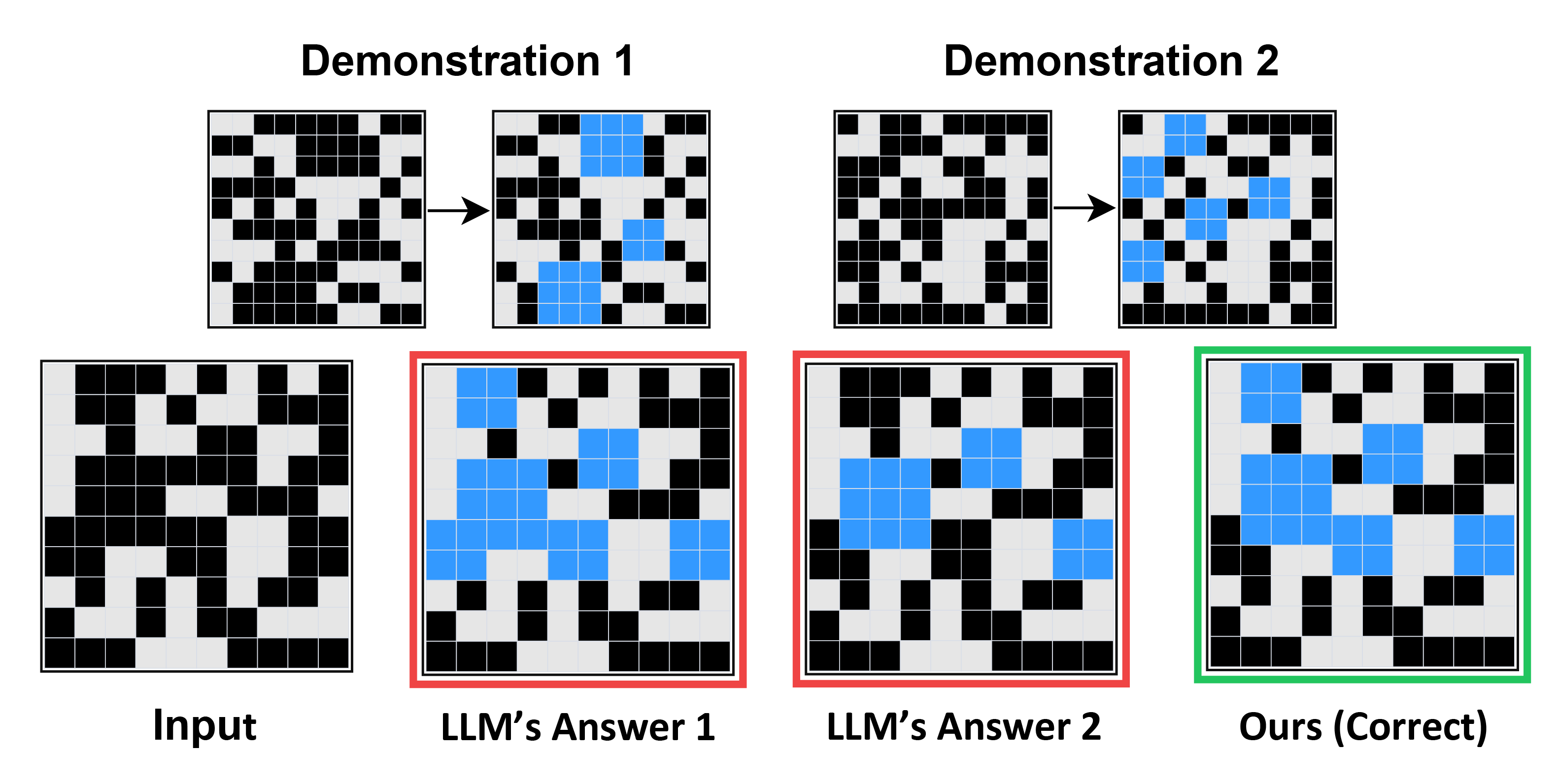}
    \caption{Case study on ARC-AGI-1 task 31adaf00.}
    \label{fig:single_case_study}
\end{figure}

Figure~\ref{fig:single_case_study} shows the prediction results of Llama-3.2-3B model on ARC-AGI-1 task 31adaf00 and see Appendix~\ref{app:case_study} for more cases. In this task, the main challenge lies in identifying all valid blank pockets rather than generating a new structure. We suggests that the output-level perturbation based negative sample is helpful when dealing with similar problems, because it preserves the global layout while removing small valid filled regions, closely matching the model’s typical under-completion errors. Thereby, our model gives the precise prediction while the original model does not.

\section{Related Work}

\paragraph{Improving ARC Reasoning of Models.}
Existing approaches to improving ARC reasoning ability can be broadly grouped into three types: (1) \textit{recursive models}~\cite{wang2025hrm,jolicoeurmartineau2025less,baek2026generativerecursivereasoning} view ARC as an iterative reasoning process in which intermediate states must be repeatedly refined, and therefore improve performance by revisiting and updating candidate solutions through repeated computation; (2) \textit{vision models} treat ARC as a visual reasoning problem~\cite{li2025tackling,hu2025arcvision,shu2026loopvitscalingvisualarc}, formulating ARC problems as an image-to-image or grid-to-grid prediction task and exploit visual inductive biases to improve performance on ARC-style benchmarks; and (3) \textit{large language models} solve ARC-style problems by redesigning task representations, prompting strategies, and ARC-oriented training data. \citet{xu2024llm4arc} shows that object-based representations can substantially improve GPT-style reasoning on ARC. Subsequent works explore explicit hypothesis generation and program verification~\cite{wang2024hypothesis,qiu2024phenomenal}, and self-improving program synthesis through hindsight replay~\cite{butt2024codeit}. More recent approaches improve ARC reasoning through data augmentation and synthetic program-based supervision \citep{bikov2024reflection,wang2025mixture}. These efforts have made some progress, but have yet to tap into the potential improvement of negative samples for ARC reasoning ability.

\paragraph{Preference Alignment of LLMs.}
Preference alignment has become a standard paradigm for adapting LLMs to desired behaviors. Early work studies learning from human preferences and reinforcement learning from human feedback, with the goal of making model outputs more helpful, honest, and harmless \citep{ouyang2022training}. More recently, direct preference optimization methods simplify this pipeline by optimizing pairwise preferences without separately training an explicit reward model \citep{rafailov2023direct}. A growing body of work has further explored simpler and more efficient preference-based objectives, such as reference-free or monolithic optimization variants \citep{hong2024orpo,meng2024simpo}. Our work extends this line of research to ARC-style reasoning by constructing task-specific chosen-rejected output pairs, so that the model learns not only to fit correct outputs, but also to distinguish them from plausible yet rule-incorrect alternatives.

\section{Conclusion}

We introduced \textsc{DiARC}, a preference alignment framework for ARC-like tasks by training LLMs to distinguish correct outputs from task-specific rejected alternatives. We proposed three negative sample construction strategies operating at different levels: output-level visual transformations, DSL-level rule inversion, and task-specific rule editing. Across six ARC-style benchmarks, our method consistently improves the ARC reasoning ability of open-source LLMs, outperforms ARC-specialized and strong closed-source models. These results suggest that, with suitable constructed negative samples, preference alignment can also be a effective post-training techniques for improving ARC-style reasoning ability of LLMs.

\section*{Limitations}

First, the full negative-construction pipeline of our method depends on auxiliary ARC resources, especially RE-ARC task generators and task-specific transformation programs. At present, such standardized per-task generation resources are available for ARC-AGI-1, but comparable RE-ARC-style generators are not yet available for the remaining benchmarks. As a result, all three negative-construction strategies can be applied on ARC-AGI-1, but only the output-level visual transformation strategy is used on other benchmarks. Second, our rejected outputs are constructed automatically rather than annotated by humans, so their informativeness and semantic quality may vary across tasks and construction methods, and the performance can be further improved when negatives are reasonably manually evaluated and assigned.

\section*{Acknowledgments}
This work was supported in parts by Guangdong Basic and Applied Basic Research Foundation (2026A1515011358), Shenzhen Natural Science Foundation (JCYJ20250604181610014), and Intelligent Computing Center of Shenzhen University.

\bibliography{custom}

\clearpage
\onecolumn
\appendix

\clearpage
\onecolumn
\appendix

\clearpage
\onecolumn
\appendix

\section{Data Statistics}
\label{app:data_statistics}

This appendix provides summary statistics of the constructed preference data used in our experiments. Table~\ref{tab:data_size_summary} reports the number of samples for output-level negative construction on the six ARC-style benchmarks, together with the additional ARC-AGI-1 data used for the DSL-level inversion and task-specific editing settings.

\begin{table}[H]
\centering
\small
\renewcommand{\arraystretch}{1.15}
\setlength{\tabcolsep}{4.5pt}
\begin{tabular}{lcccccccc}
\toprule
\multirow{2}{*}{\textbf{Metric}} & \multicolumn{6}{c}{\textbf{Output-Level}} & \textbf{DSL-Level} & \textbf{Task-Specific} \\
\cmidrule(lr){2-7} \cmidrule(lr){8-9}
& \textbf{ARC-1} & \textbf{ARC-2} & \textbf{MiniARC} & \textbf{Concept} & \textbf{1D} & \textbf{Community} & \multicolumn{2}{c}{\textbf{ARC-1}} \\
\midrule
\textbf{\# Samples} & 15,000 & 12,000 & 20,000 & 12,000 & 12,000 & 2,000 & 6,000 & 4,000 \\
\bottomrule
\end{tabular}
\caption{Summary of data quantities used in our experiments.}
\label{tab:data_size_summary}
\end{table}

\section{Additional Training Details}
\label{app:training_details}

\begin{table}[h]
\centering
\small
\setlength{\tabcolsep}{8pt}
\renewcommand{\arraystretch}{1.18}
\rowcolors{3}{gray!10}{white}
\begin{tabularx}{0.98\textwidth}{>{\raggedright\arraybackslash}p{0.28\textwidth}YYY}
\toprule
& \textbf{Qwen3-4B} & \textbf{Llama-3.2-3B} & \textbf{Minitron-8B} \\
\midrule
DPO $\beta$
& 0.1 & 0.1 & 0.1 \\

Learning rate
& $1\times10^{-6}$ & $1\times10^{-6}$ & $1\times10^{-6}$ \\

Epochs
& 1 & 1 & 1 \\

Per-device batch size
& 2 & 1 & 2 \\

Gradient accumulation steps
& 4 & 8 & 4 \\

Effective batch size
& 8 & 8 & 8 \\

LoRA rank $r$
& 256 & 32 & 32 \\

LoRA $\alpha$
& 32 & 16 & 16 \\

LoRA dropout
& 0.05 & 0.05 & 0.05 \\

Precision / loading
& bf16; fp16 fallback
& NF4 4-bit + bf16
& NF4 4-bit + bf16 \\

Optimizer
& \texttt{paged\_adamw\_8bit}
& \texttt{paged\_adamw\_8bit}
& \texttt{paged\_adamw\_8bit} \\

Scheduler
& cosine & cosine & cosine \\

Warmup ratio
& 0.1 & 0.1 & 0.1 \\

Max sequence length
& 4096 & 4096 & 4096 \\

Max prompt length
& 3584 & 3584 & 3584 \\

Hardware
& 1x NVIDIA L40 & 1x NVIDIA L40 & 1x NVIDIA L40 \\
\bottomrule
\end{tabularx}
\caption{Detailed training settings for the DPO-based instantiation of \textsc{DiARC}. For all models, LoRA is applied to \texttt{q\_proj}, \texttt{k\_proj}, \texttt{v\_proj}, \texttt{o\_proj}, \texttt{gate\_proj}, \texttt{up\_proj}, \texttt{down\_proj}, \texttt{embed\_tokens}, and \texttt{lm\_head}.}
\label{tab:training_details}
\end{table}

\section{Case Study}
\label{app:case_study}

\begin{figure}[H]
    \centering
    \includegraphics[width=.98\linewidth]{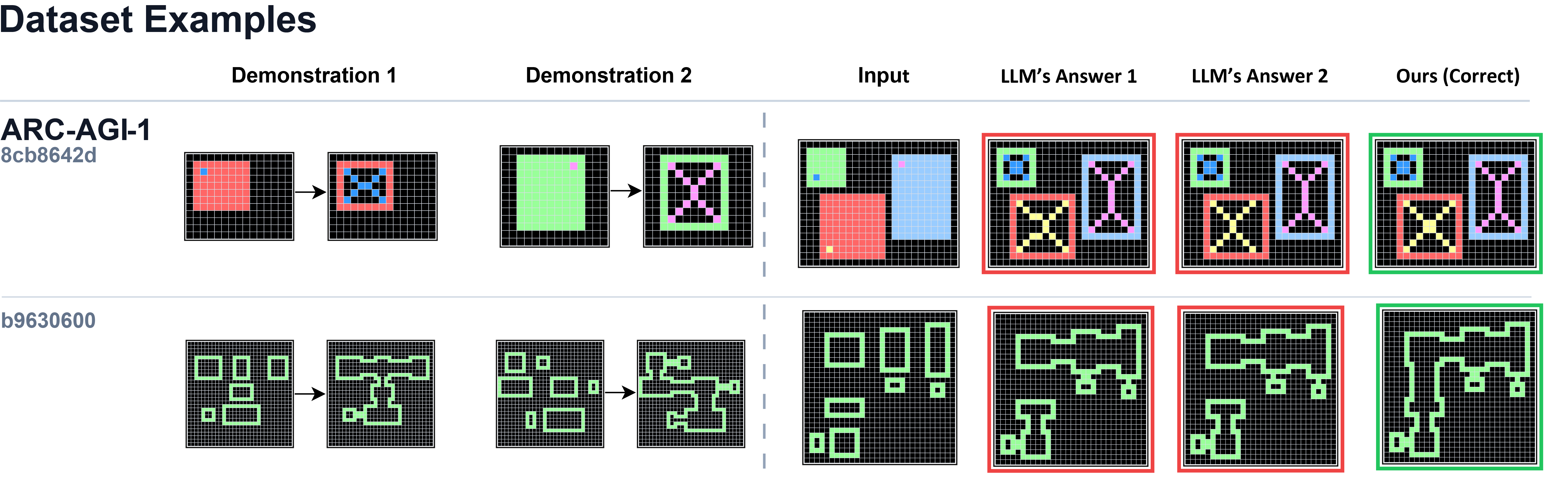}
    \caption{Cases on ARC-AGI-1, where the base model fails while we predict correctly.}
    \label{fig:case_arc12}
\end{figure}

\begin{figure}[H]
    \centering
    \includegraphics[width=.98\linewidth]{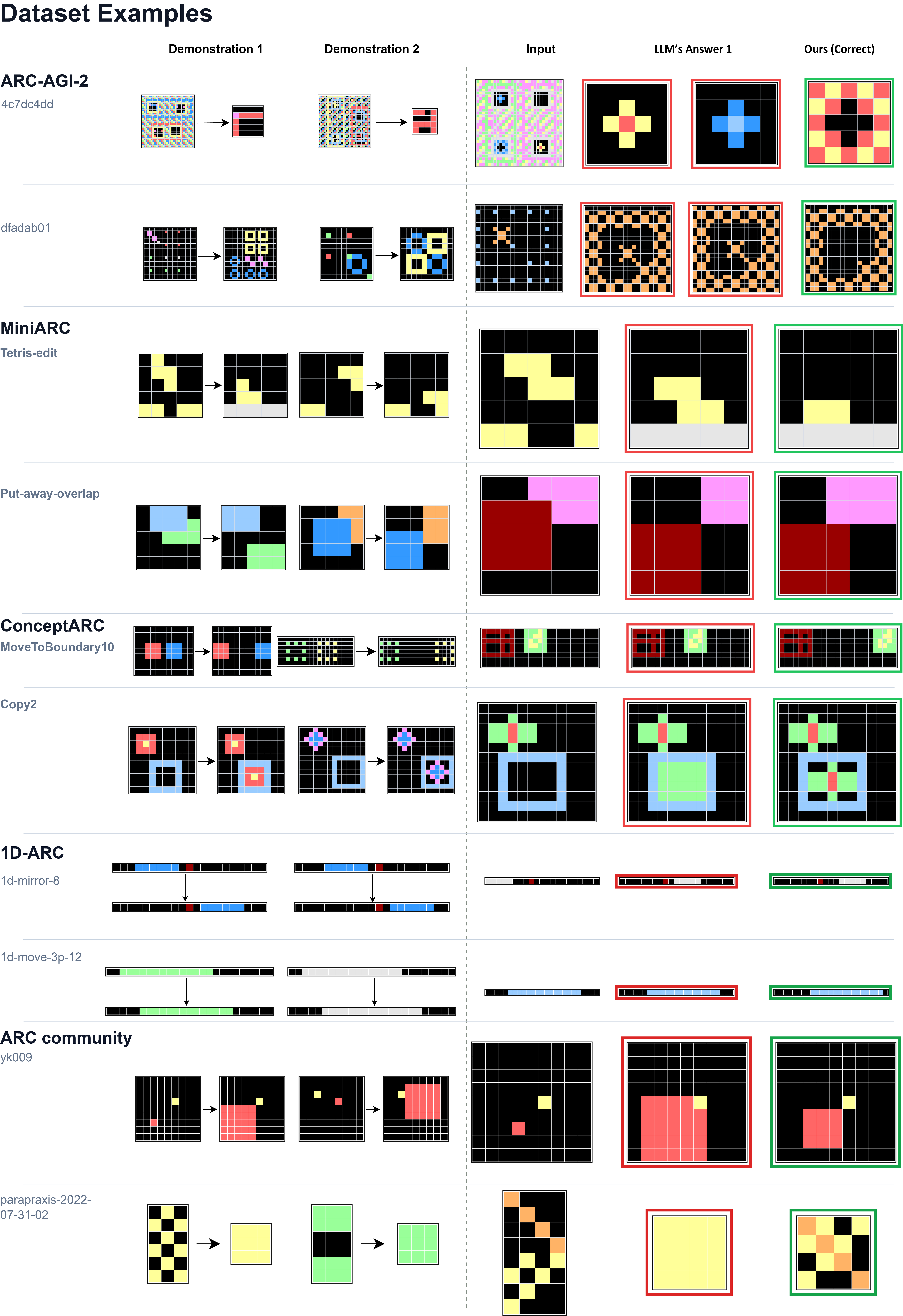}
    \caption{Cases on ARC-AGI-2,MiniARC, ConceptARC, 1D-ARC, and ARCcommunity, where the base model fails while we predict correctly.}
    \label{fig:case_transfer}
\end{figure}

We present more representative cases on ARC-AGI-1 in Figure~\ref{fig:case_arc12}, and on ARC-AGI-2 ,MiniARC, ConceptARC, 1D-ARC, and ARCcommunity in Figure~\ref{fig:case_transfer}, respectively.

\section{Additional Visualizations of Negative Construction Strategies}
\label{app:extra_ops}

We provide additional visualizations of the negative-construction operations that are not individually shown in the main pages.

\subsection{Output-Level Visual Transformations}
\label{app:extra_output}

\begin{figure}[H]
    \centering
    \includegraphics[width=.98\linewidth]{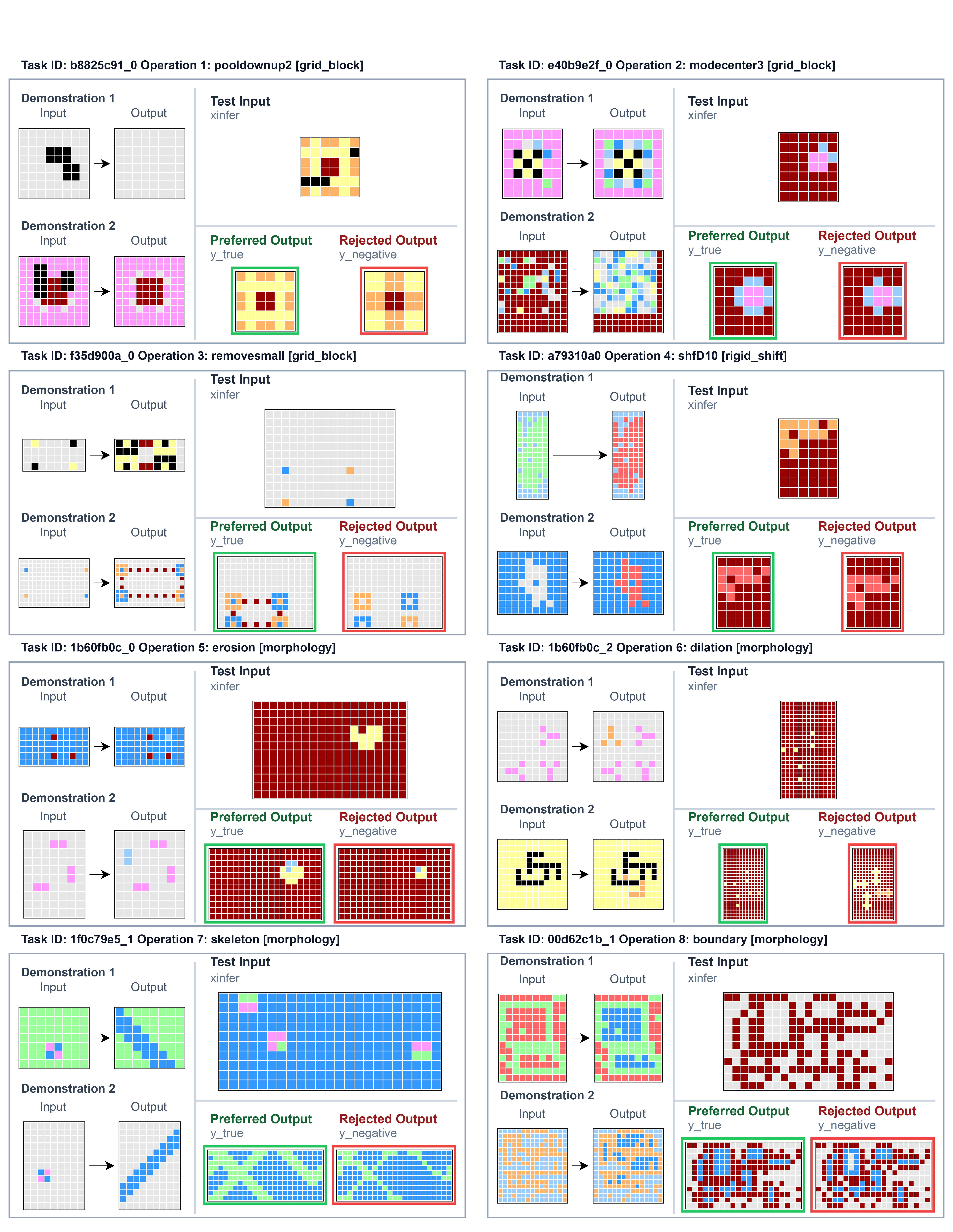}
    \caption{Additional examples of output-level visual transformations.}
    \label{fig:extra_output_1}
\end{figure}

\begin{figure}[H]
    \centering
    \includegraphics[width=.98\linewidth]{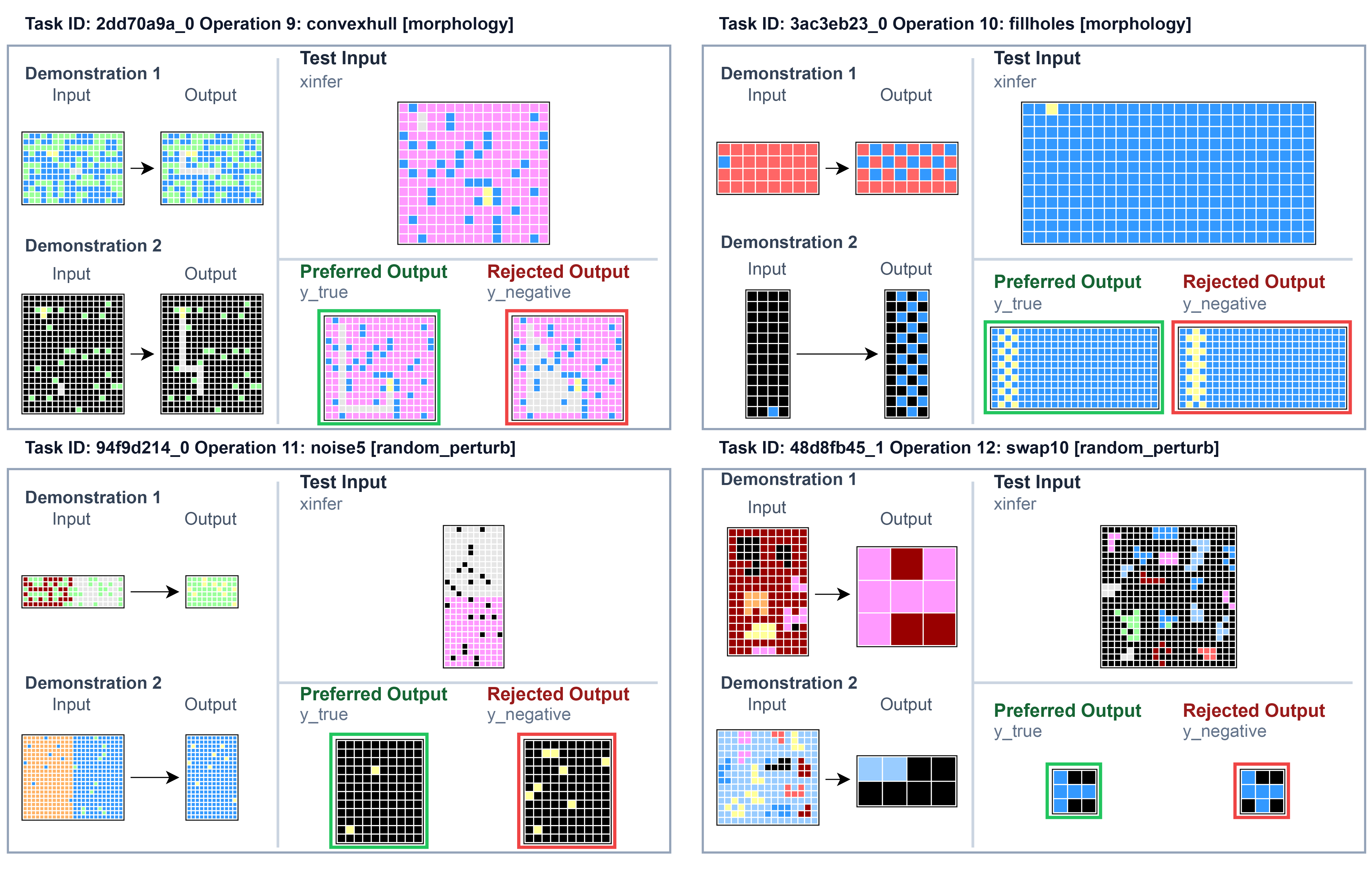}
    \caption{Additional examples of output-level visual transformations.}
    \label{fig:extra_output_2}
\end{figure}

Figures~\ref{fig:extra_output_1} and~\ref{fig:extra_output_2} show additional examples of output-level visual transformations.

\subsection{DSL-Level Rule Inversion}
\label{app:extra_dsl}

\begin{figure}[H]
    \centering
    \includegraphics[width=.98\linewidth]{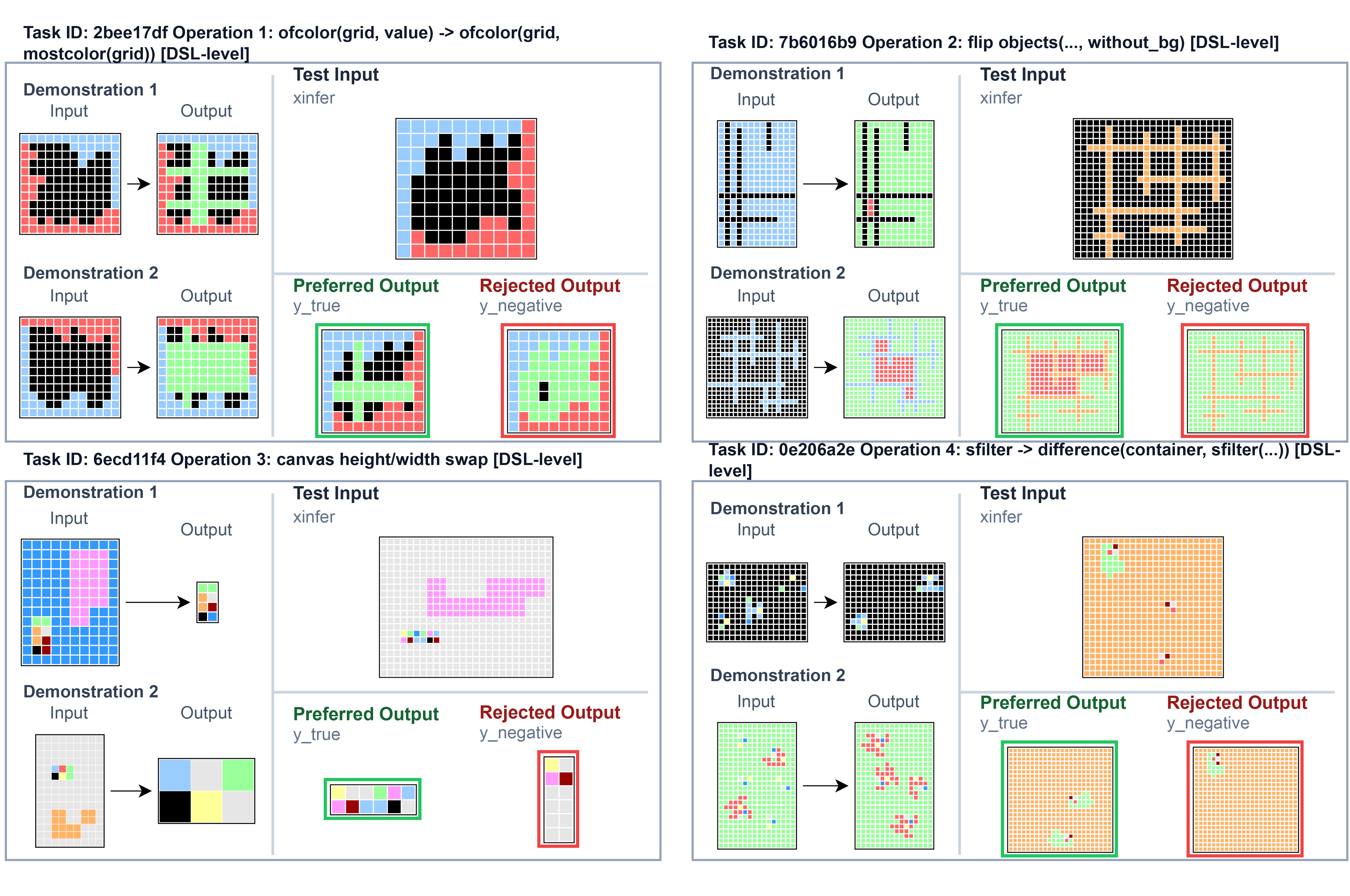}
    \caption{Additional examples of DSL-level rule inversion.}
    \label{fig:extra_dsl_1}
\end{figure}

\begin{figure}[H]
    \centering
    \includegraphics[width=.98\linewidth]{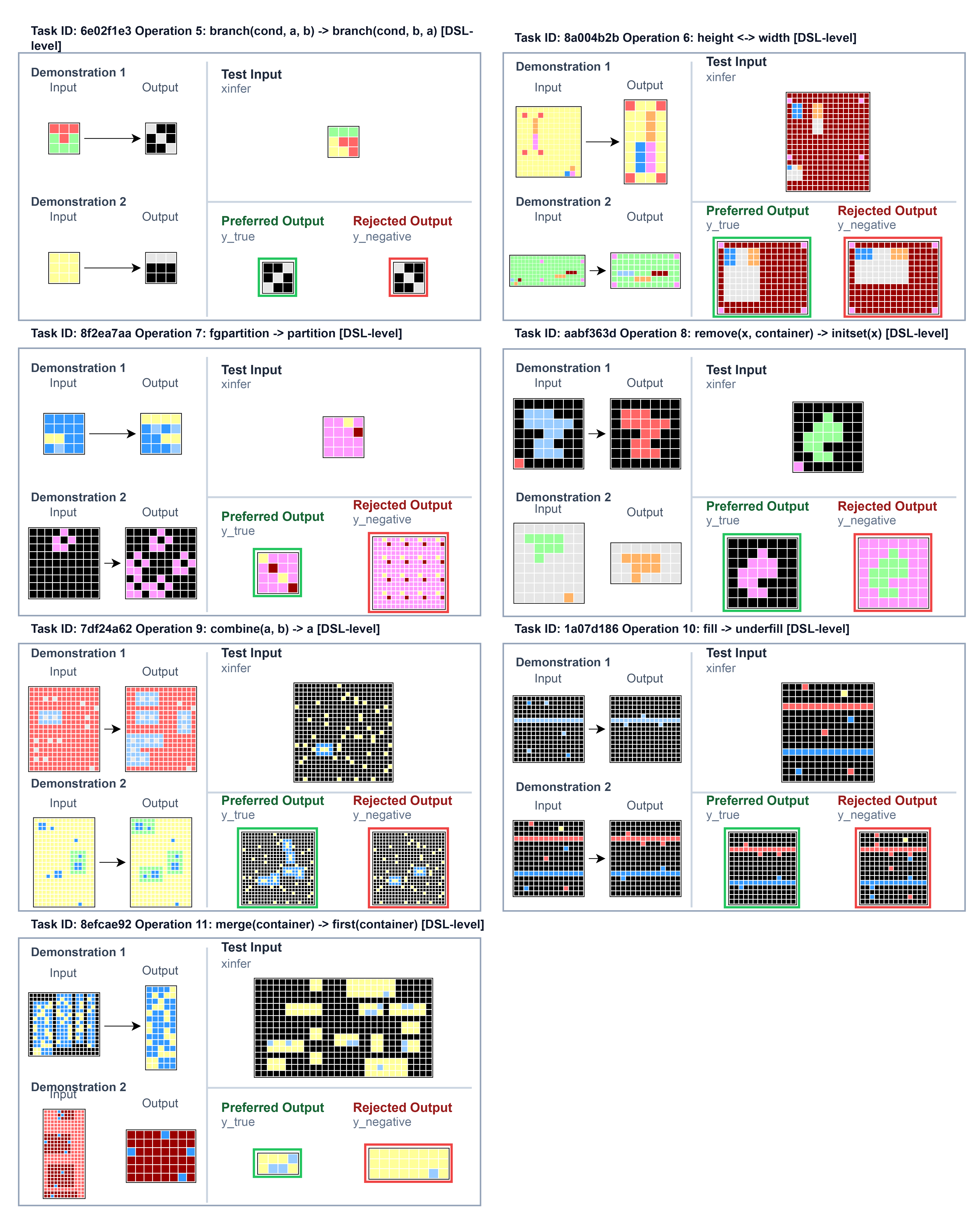}
    \caption{Additional examples of DSL-level rule inversion.}
    \label{fig:extra_dsl_2}
\end{figure}

Figures~\ref{fig:extra_dsl_1} and~\ref{fig:extra_dsl_2} show additional examples of DSL-level rule inversion.

\subsection{Task-Specific Rule Editing}
\label{app:extra_edit_sec}

\begin{figure}[H]
    \centering
    \includegraphics[width=.98\linewidth]{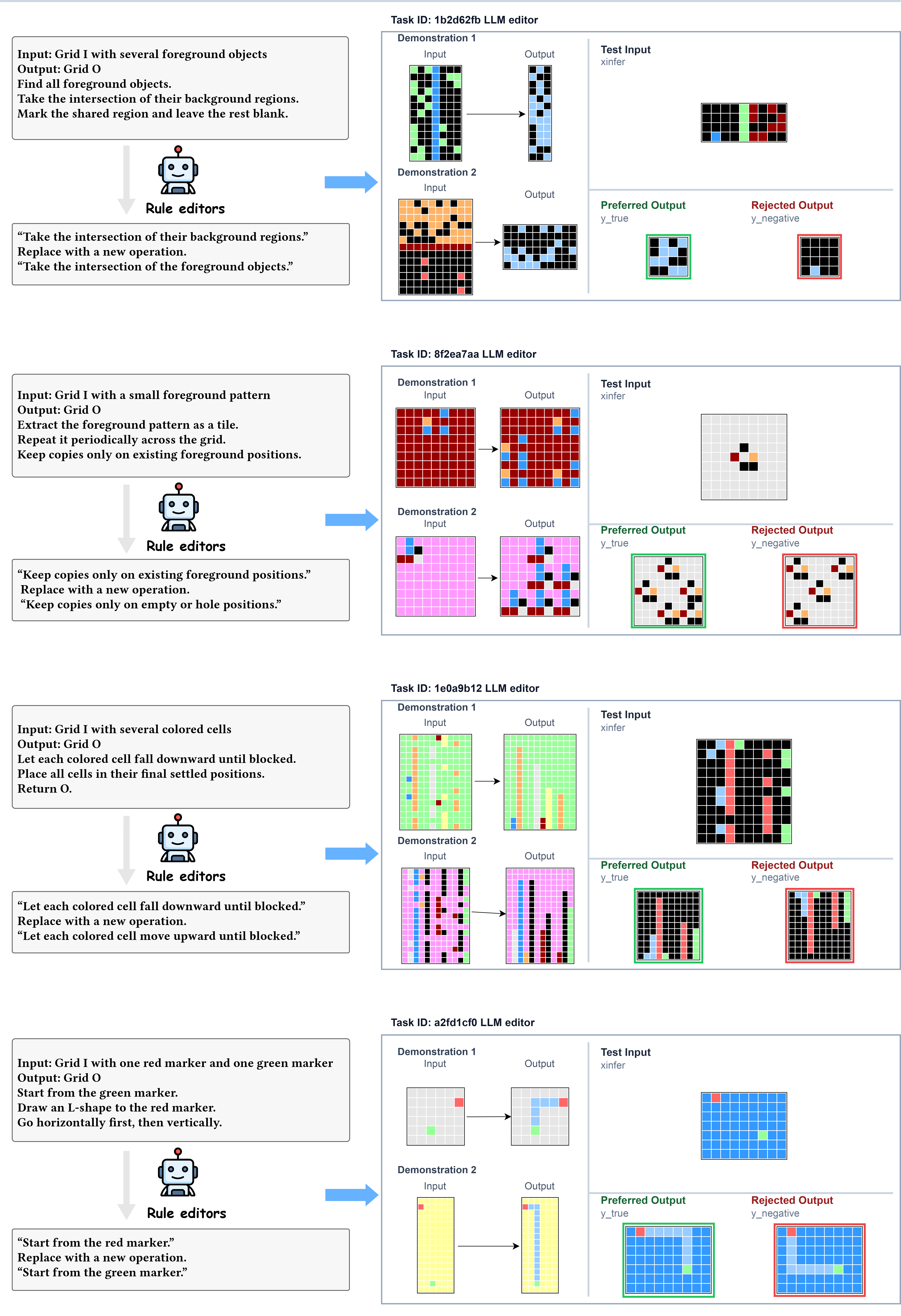}
    \caption{Additional examples of task-specific rule editing.}
    \label{fig:extra_edit}
\end{figure}

\begin{figure*}[t]
    \centering
    \includegraphics[width=.9\linewidth]{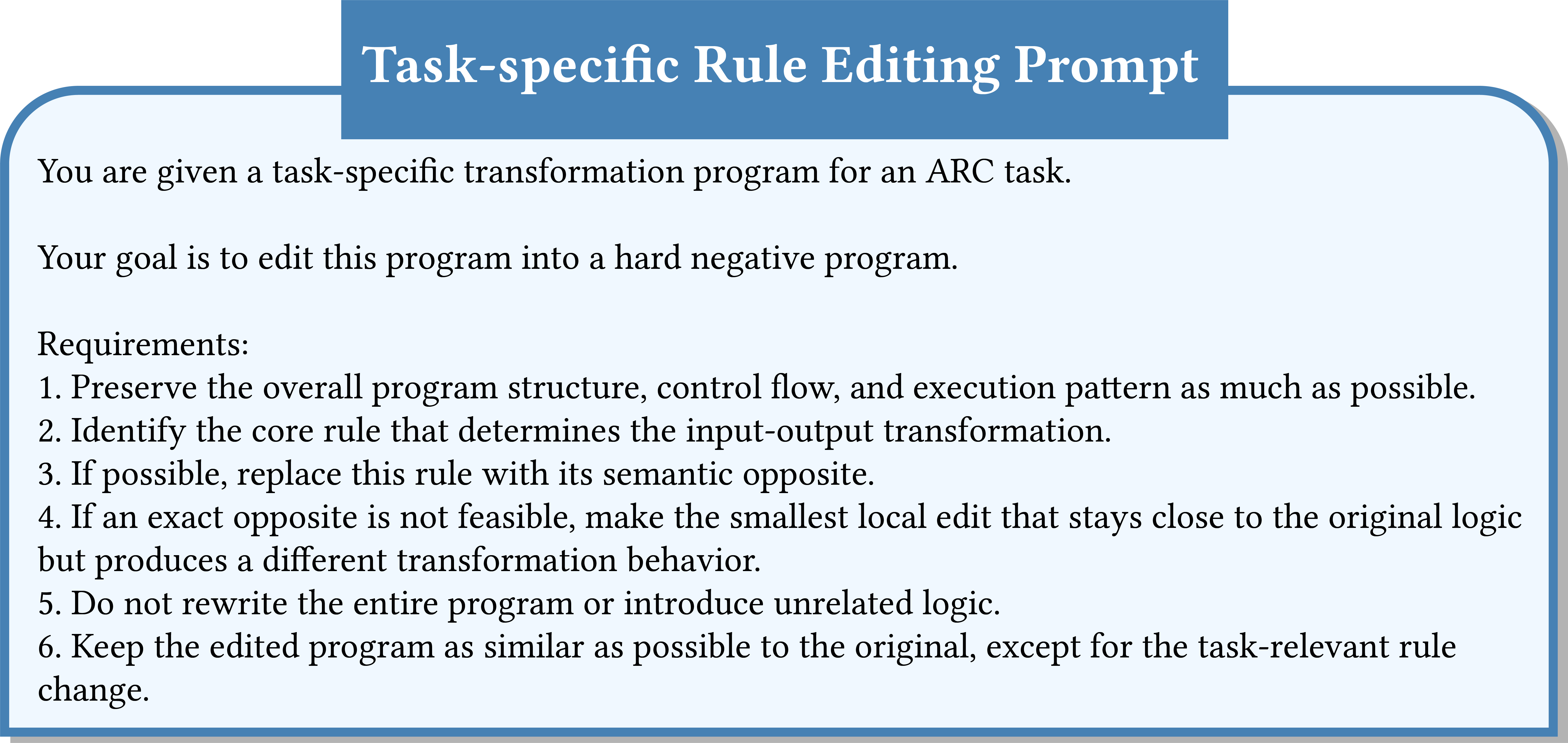}
    \caption{Prompt template used for task-specific rule editing with GPT-5.4.}
    \label{fig:prompt}
\end{figure*}

Figure~\ref{fig:extra_edit} shows additional examples of task-specific rule editing. We used the prompt template in Figure~\ref{fig:prompt} to guide GPT-5.4 in editing task-specific transformation rules and constructing negative samples.

\section{AI Assistance Statement}
\label{app:ai_assistance}

We utilized GPT-5.4 to assist in polishing the text, refining grammar, and improving the clarity of the manuscript. All content was reviewed and verified by the authors. We also employed GPT-5.4 in the task-specific rule editing process described in Appendix~\ref{app:extra_edit_sec}. The generated content was reviewed and verified by the authors.

\end{document}